# Compressed sensing MRI using masked DCT and DFT measurements

Elma Hot, Petar Sekulić
Faculty of Electrical Engineering
University of Montenegro
Podgorica, Montenegro
elma_hot@live.com, petar77@live.com

*Abstract*—**This paper presents modification of the TwIST algorithm for Compressive Sensing MRI images reconstruction. Compressive Sensing is new approach in signal processing whose basic idea is recovering signal form small set of available samples. The application of the Compressive Sensing in biomedical imaging has found great importance. It allows significant lowering of the acquisition time, and therefore, save the patient from the negative impact of the MR apparatus. TwIST is commonly used algorithm for 2D signals reconstruction using Compressive Sensing principle. It is based on the Total Variation minimization. Standard version of the TwIST uses masked 2D Discrete Fourier Transform coefficients as Compressive Sensing measurements. In this paper, different masks and different transformation domains for coefficients selection are tested. Certain percent of the measurements is used from the mask, as well as small number of coefficients outside the mask. Comparative analysis using 2D DFT and 2D DCT coefficients, with different mask shapes is performed. The theory is proved with experimental results.**

*Keywords-component; Compressed sensing, CS, Magnetic resonance imaging, MRI*

## I. Introduction

Compressed sensing (CS) is a novel signal processing technique. It aims to reconstruct signals from reduced number of measurements, significantly lowering number of samples for reconstruction, compared the number required by Shannon-Nyquist theorem [1]-[7]. CS assumes that signal satisfies sparsity property. Also, CS requires incoherent acquisition procedure. Sparsity states that there must be a domain in which whole signal energy is concentrated within small number of coefficients in that domain. These coefficients have values significantly larger than others. For such signals, processing and analysis can be done using small number of coefficients from the domain where signal is dense. These signals are considered sparse and that is the basis condition for successful CS approach. Incoherence property is satisfied if the samples from dense domain are acquired in certain way. It is shown that, by random selection of samples for CS, incoherence property will be satisfied. This property is important to assure successful reconstruction from small set of acquired samples. CS reconstruction is based on minimization different signal norms. Common is minimization of the $l_1$ norm, and in that sense, different optimization algorithms are used [8]-[11].

A majority of real signals satisfy these two conditions and therefore, we might say that the field of CS application is large. CS is used in a mobile phone camera sensor, in facial recognition applications and commercial shortwave-infrared cameras (etc). It also can be used to improve image reconstruction in holography and to reduce magnetic resonance imaging (MRI) scanning sessions [12]-[17]. MRI is a medical imaging technique which uses strong magnetic fields and radio waves to produce images of human body. MRI is based on the movement of the protons of the hydrogen nuclei, which comprises a spin or a magnetic moment. The human body, for the most part, made up of fat and water, respectively 63% of the hydrogen atoms. When the patient is exposed to a strong magnetic field, all of its protons agree in the direction of the magnetic field. As a result of rotation of the proton about the level of a strong magnetic field induces an electric current, the MR signal, which registered coil. MRI scans can be unpleasant for those who are claustrophobic, for babies and other young children, patients with cochlear implants and cardiac pacemakers, shrapnel and metallic foreign bodies in patient body. That means imaging speed is important in MRI applications. Use of CS in MRI offers significant scan time reduction.

Some MR images such as angiograms are already sparse in the pixel representation, while other MR images are more complicated but have a sparse representation in some transform domain. As majority of 2D signals are not sparse nor in space neither in frequency domain, minimization is done on the image gradient, which is sparse signal. In this paper, CS measurements are collected in frequency domain. We have observed 2D discrete cosine transform (DCT) and 2D discrete Fourier transform (DFT) domains. The frequency samples are used form the a priori defined masks. Different mask shapes are observed as well. Certain percent of the samples is used form the mask. We have tested case when the certain number of samples outside the mask is used together with the samples form the mask. The reconstruction quality is measured with peak signal to noise ratio (PSNR).

The paper is organized as follows: The theoretical part on CS is given in Section II. Section III describes standard TwIST

algorithm and proposed modification. Experimental results and concluding remarks are given in Sections IV and V, respectively.

## II. COMPRESSIVE SAMPLING

One-dimensional signal *x* can be shown as a column vector of length *N*, i.e., in terms of basis vectors [1]-[3]:

$$x = \sum_{i=1}^{N} \mu_i \beta_i = \beta\mu, \quad (1)$$

where β is matrix of transformation domain coefficients and μ is transformation vector, ie., it is representation of signal *x* in β domain. Dimension of β is *N*×*N*. Having in mind the first CS requirement for the successful reconstruction, sparsity, we might say that signal *x* has to be sparse in domain β. The second requirement is incoherence, which is related to the measurement procedure - signal measurement procedure should be incoherent. Incoherence defines how many samples we will need at least to allow reconstruction from small number of samples. We will assume random selection of the coefficients, to assure incoherence property to be valid.

Number of collected samples of signal or measurements of is *M*, *M*<<*N*. We form measurement vector *y* of length *M* by multiplying the measurement matrix *F* with the vector *x*.

$$y = Fx. \quad (2)$$

That means:

$$y = Fx = F\beta\mu \quad (3)$$

We have *y* samples from which signal will be reconstructed. This reconstruction involves complex calculations. Complicated mathematical algorithms are used for realization of CS. In our experiment we have used " A New TwIST: Two-Step Iterative Shrinkage thresholding [12].

## III. ALGORITHMS FOR IMAGE RECONSTRUCTION

### A. Iterative shrinkage/thresholding (IST)

Iterative shrinkage/thresholding (IST) [12] algorithms handle with optimization problems arising in image restoration and other linear inverse problems. For finite-dimensional case:

$$X = R^m, \quad (4)$$
$$Y = R^n. \quad (5)$$

IST algorithms has the form:

$$x_{t-1} = (1-B)x_t + BF_\lambda(x_t + K^T(y - Kx_t)) \quad (6)$$

Each iteration of the IST algorithm only involves sums, matrix-vector multiplication by *K* and $K^T$, and the application of the denoising operation $F_\lambda$. The original IST algorithm has the form (6) with *B*=1.
Convergence rate of these IST algorithms depends on the linear observation operator and can be very slow in the cases of ill-conditioned or ill-posed linear observation operator.

When is strongly ill-conditioned then iterative reweighted shrinkage (IRS) algorithm much faster than IST. But IST is faster than IRS for mildly ill-conditioned and medium to strong noise [12].

### B. A New TwIST: Two-Step Iterative Shrinkage thresholding(TwIST)

To solve this speed problem we use new class of iterative schemes, which brings together the best of IRS and IST. Algorithms in this class have a two-step IST (TwIST) [12] structure, each iterate depends on the two previous iterates, rather than only on the previous one.

$$C = I + \lambda D_t, \quad (7)$$

where $D_t$ is a diagonal matrix of non-negative elements. $D_t$ depends on *x* and ϕ. ϕ is a function $\phi: X \to \overline{R}$ and it is called the regularizer. λ is the regularization parameter, $\lambda \in [0, +\infty]$.

$$R = I - K^T K, \quad (8)$$
$$A = C - R, \quad A = \lambda D_t + K^T K, \quad (9)$$
$$Ax = K^T y. \quad (10)$$

For the two-step iteration for the linear system becomes:

$$x_{t+1} = (1-a)x_{t-1} + (a-B)x_t + BC^{-1}(x_t + K^T(y - Kx_t)). \quad (11)$$

Setting *a*=1 and replacing the multiplication $C^{-1}$ matrix by the denoising operator $F_\lambda$. Two-step version of IST (TwIST) is defined as:

$$x_1 = \Gamma_\lambda(x_0), \quad (12)$$
$$X_{t+1} = (1-a)x_{t-1} + (a-B)x_t + B\Gamma_\lambda(x_t), \quad (13)$$
$$\Gamma_\lambda : R^m \to R^m, \quad \Gamma_\lambda = F_\lambda(x + K^T(y - Kx)), \quad (14)$$

Convergence rate of TwIST algorithms is much faster than IST for ill-conditioned problems. The TwIST method, which we used in experiments, keeps the good denoising performance of the IST scheme, while it is still able to handle ill-posed problems as good as the IRS algorithm. When *B*=1 TwIST, IST and original IST have the same fixed points.

$$x = (1-a)x + (a-B)x + B\Gamma_\lambda(x)$$
$$x = (1-B)x + B\Gamma_\lambda(x), \quad x = \Gamma_\lambda(x) \quad (15)$$

### C. Proposed modification of the TwIST algorithm

We used "A New TwIST: Two-Step Iterative Shrinkage thresholding" algorithms for image restoration. First we transform MR image (Figure 1) from time domain to frequency domain. We have used MATLAB functions FFT2 and DCT2 for 2D DFT and 2D DCT, respectively. Different mask shapes are formed, depending on the energy distribution. When observing 2D DFT, energy of the image is concentrated around the origin. Therefore, we have used masks around the origin to collect samples. We used the square and circular mask shapes. Square mask is made according to the equation:

$$y(n_1, n_2) = \begin{cases} 1, \dfrac{N}{2} - 0{,}1*N \le n_1 \le \dfrac{N}{2} + 0{,}1*N \\ \quad \wedge \ \dfrac{N}{2} - 0{,}1*N \le n_2 \le \dfrac{N}{2} + 0{,}1*N \\ 0, \text{ for other values } n_1 \text{ and } n_2 \end{cases} \quad (16)$$

Circular mask is made according to the equation:

$$y(n_1, n_2) = \begin{cases} 1, \left(n_1 - a\dfrac{N}{2}\right)^2 + \left(n_2 - a\dfrac{N}{2}\right)^2 \le \left(\dfrac{N}{b}\right)^2 \\ 0, \text{ for other values } n_1 \text{ and } n_2 \end{cases} \quad (17)$$

Parameter *a* is used to choose right position of center of the mask. Parameter *b* is used to adjust radius of the circular mask. The second mask is circular mask with the center in the middle of the image. For that mask, in the experiments, we have used relation (17) with parameters $a=1$, $b=6$. The radius of this circle is $(N/4)$. Mask with the center in the origin is also created by using relation (17) and parameters *a* and *b* equal to 0 and 2, respectively. We have chosen radius of this circle to be $(2*N/3)$ in experimental results part, for this mask type. From the image in frequency domain, we take only a certain percentage of random samples from the mask and around it. The largest part of the energy picture is located in the center of the FFT image but for DCT image most of energy is in the upper left part of the 2D DCT plane. Above-mentioned algorithm gave different results for same image in different domains and with different masks. Analyzing image quality we have done comparative analysis using different domains.

## IV. EXPERIMENTAL RESULTS

The experiments were done on MRI image of size 512×512 and by using 2D DCT and 2D DFT domains with different masks. We tested TwIST algorithm taking a certain percentage of samples from the mask, and a smaller part around the masks. On this basis, we perform image reconstruction. Original image is shown in Fig. 1.

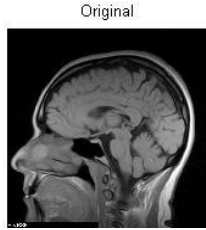

Figure 1: Original image MRI Brain

Figs 2 and 3 show reconstruction using different masks in DFT domain. Fig. 2 shows reconstruction with square mask, when 80% samples form the mask and 20% samples around the mask is used (Fig. 2b) and reconstruction when 100% samples from the mask and 10% samples around the mask is used (Fig. 2c). Fig. 3 shows reconstruction with same percent of the samples form and around the mask, but in this case circular mask is used.

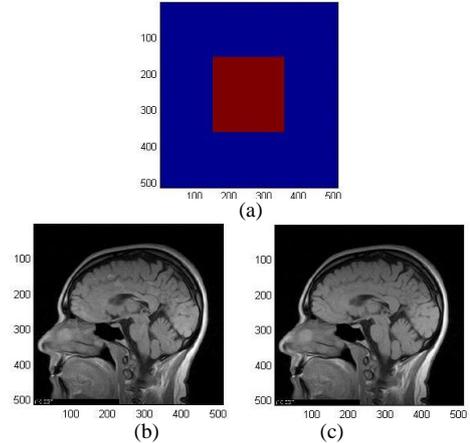

Figure 2: a) Mask b) estimate -80% from mask, 20% around mask
c) estimate-100% from mask,10% around mask

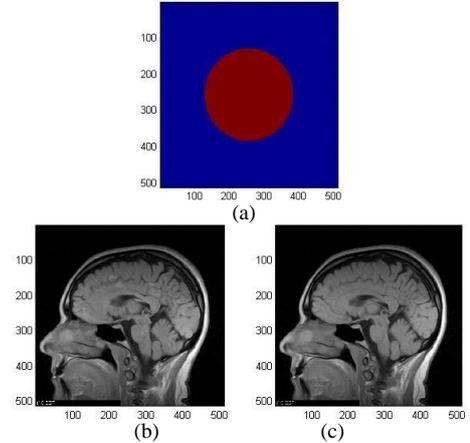

Figure 3: a)Mask b)estimate -80% from maske, 20% around mask
c) estimate-100% from mask,10% around mask

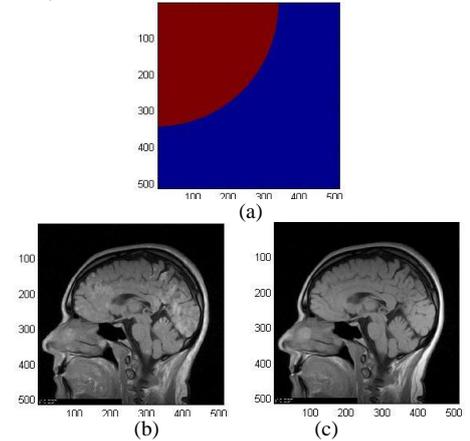

Figure 4: a)Mask b)estimate -80% from maske, 20% around mask
c) estimate-100% from mask,10% around mask

In the case of 2D DCT transform domain, mask is formed in left upper corner of the 2D DCT plane (Fig. 4a). The reconstruction is done using the same number of coefficients: first, 80% samples form the mask and 20% samples around the mask (Fig. 4b), and second 100% samples from the mask and 10% samples around the mask (Fig. 4c). We compared the PSNR for different cases. Fig. 5 shows PSNR as a function of percentage of used measurements. PSNR in Fig. 5a is calculated using the square mask as it is shown in Fig. 2a, while PSNR in Fig. 5b is calculated using the circular mask as it is shown in Fig. 3a. We used whole mask and different percentage of samples around mask. PSNR dependence on the percentage of measurements, when 2D DCT domain is considered, is shown in Fig. 6.

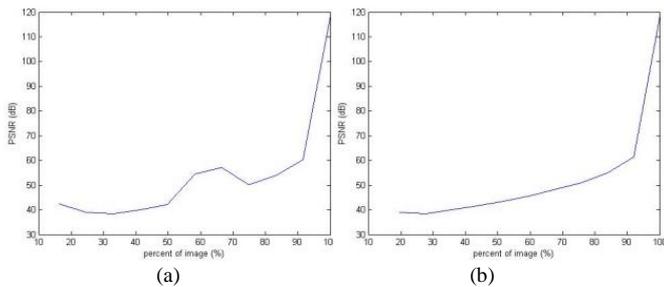

(a)                                (b)

Figure 5: PSNR as a function of the percentage of image, FFT (a) square mask (b) circular mask

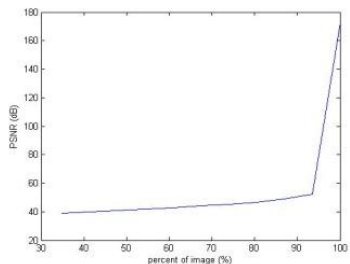

Figure 6 : PSNR as a function of the percentage of image, DCT, circular mask with the center at the origin

## V. CONCLUSION

A modification of the commonly used TwIST algorithm for MRI images reconstruction is presented in the paper. Standard TwIST algorithm chooses CS measurement from the 2D FFT domain. Samples from the 2D FFT are selected by using radial line mask. In this paper, other mask forms are observed as well as other frequency domain for measurements selection – 2D DCT domain. Having in mind that the energy in the 2D FFT is concentrated around the origin, square and round masks in the central part of the 2D FFT plane are considered. Regarding the 2D DCT domain, mask is created in such way which collects certain number of high frequency DCT coefficients. For all mask types we have used certain number of samples form the mask and certain number outside the mask. The quality of reconstruction is measured with PSNR, for different masks and different percent of the used image coefficients.

## ACKNOWLEDGEMENT

The authors are thankful to Professors and assistants within the Laboratory for Multimedia Signals and Systems, at the University of Montenegro, for providing the ideas, codes, literature and results developed for the project CS-ICT (funded by the Montenegrin Ministry of Science).